\documentclass[journal]{IEEEtran}
\usepackage{cite}

\ifCLASSINFOpdf
   \usepackage[pdftex]{graphicx}
   \graphicspath{{../pdf/}{../jpeg/}}
   \DeclareGraphicsExtensions{.pdf,.jpeg,.png}
\else
   \usepackage[dvips]{graphicx}
   \graphicspath{{../eps/}}
   \DeclareGraphicsExtensions{.eps}
\fi
\usepackage{multirow}
\usepackage{color}
\usepackage{float}

\usepackage{amsmath,amssymb}
\usepackage{algorithmic}

\usepackage{array}

\ifCLASSOPTIONcompsoc
 \usepackage[caption=false,font=normalsize,labelfont=sf,textfont=sf]{subfig}
\else
 \usepackage[caption=false,font=footnotesize]{subfig}
\fi
\usepackage{url}

\begin{document}
%
\title{Spatiotemporal Dilated Convolution with Uncertain Matching for Video-based Crowd Estimation}
%
%
\author{Yu-Jen Ma, Hong-Han Shuai,\IEEEmembership{~Member,~IEEE}, and Wen-Huang Cheng,\IEEEmembership{~Senior Member,~IEEE}
\IEEEcompsocitemizethanks{
\IEEEcompsocthanksitem Y.-J. Ma and H.-H. Shuai are with the Department of Electrical and Computer Engineering, National Chiao Tung University, Hsinchu, Taiwan. E-mail: \{ninechi.eed08g,hhshuai\}@nctu.edu.tw.
\IEEEcompsocthanksitem W.-H.~Cheng is with the Institute of Electronics, National Chiao Tung Univresity, Hsinchu, 300 Taiwan, and the Artificial Intelligence and Data Science Program, National Chung Hsing University, Taichung, 400 Taiwan. E-mail: whcheng@nctu.edu.tw.
}}

\maketitle

\begin{abstract}
In this paper, we propose a novel SpatioTemporal convolutional Dense Network (STDNet) to address the video-based crowd counting problem, which contains the decomposition of 3D convolution and the 3D spatiotemporal dilated dense convolution to alleviate the rapid growth of the model size caused by the Conv3D layer. Moreover, since the dilated convolution extracts the multiscale features, we combine the dilated convolution with the channel attention block to enhance the feature representations. Due to the error that occurs from the difficulty of labeling crowds, especially for videos, imprecise or standard-inconsistent labels may lead to poor convergence for the model. To address this issue, we further propose a new patch-wise regression loss (PRL) to improve the original pixel-wise loss. Experimental results on three video-based benchmarks, i.e., the UCSD, Mall and WorldExpo'10 datasets, show that STDNet outperforms both image- and video-based state-of-the-art methods. The source codes are released at \url{https://github.com/STDNet/STDNet}.
\end{abstract}


%
\IEEEpeerreviewmaketitle

\section{Introduction}
\label{Intro}
%
%
%
%
\IEEEPARstart{C}{rowd} counting plays an important role in computer vision since it facilitates a variety of fundamental applications, \emph{e.g.,} automatic driving technologies~\cite{basalamah2016automatic,car1,car2}, video surveillance~\cite{saxena2008crowd,zhang2015crowd}, safety management~\cite{ihaddadene2008real,xu2017efficient,safety1}, and tracking~\cite{tracking1,tracking2,tracking3}. The goal is to estimate the accurate number of people in a particular region from a video or an image. The early studies of crowd counting were based on detection frameworks with well-trained classifiers~\cite{dollar2011pedestrian,subburaman2012counting,viola2005detecting}. However, crowd counting is still a challenging task due to multiple issues, such as scale variations, perspective distortion, people occlusion, and light illumination.

In recent years, benefiting from the powerful capability of feature extraction by deep convolutional neural networks (CNNs)~\cite{he2016deep,krizhevsky2012imagenet,simonyan2014very,deep15}, many CNN-based methods~\cite{cao2018scale,dai2019dense,li2018csrnet,liu2018decidenet,liu2019context,liu2019point,ranjan2018iterative,sam2017switching,sindagi2017generating,tian2019padnet,zhang2016single,zou2019enhanced} have been developed to improve the performance on crowd counting benchmarks. However, most existing approaches, even if video datasets are available, still exploit image-based methods for crowd counting while ignoring the rich information in the temporal domain. In contrast, a recent line of research attempts to leverage the temporal correlation to improve the model~\cite{zou2019enhanced,xiong2017spatiotemporal,fang2019locality,wu2020triple}. For example, \cite{xiong2017spatiotemporal} proposed the ConvLSTM to incorporate the convolutional operation into the LSTM cell to capture the temporal information, while E3D~\cite{zou2019enhanced} integrated the 3D convolutional layer into the model to jointly consider the temporal and spatial dimensions. Moreover,~\cite{fang2019locality} proposed the Locality-constrained Spatial Transformer (LSTN) module to improve the density map of the current frame by considering the information of its previous frame. TACCN~\cite{wu2020triple} proposed three complementary attention mechanisms to achieve the more robust estimation.


\begin{figure}[!t]
\centerline{\includegraphics[width=\columnwidth]{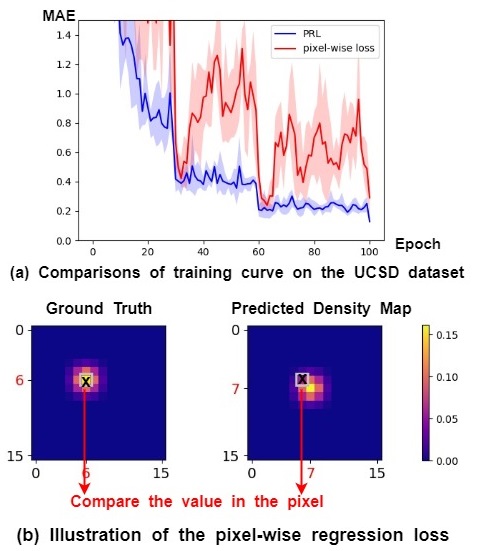}}
\caption{An illustrative example of the problem of pixel-wise regression loss. (a) shows that the pixel-wise loss has a larger oscillation scale than the proposed Patch-wise Regression Loss. The reason may be attributed to (b), \emph{i.e.}, the pixel-wise loss is sensitive to such pixel-level difference between two Gaussian blobs.}
\label{Intro_illus}
\end{figure}


Despite their comparable results, such models incorporated with temporal information still suffer from three issues. First, to capture the temporal information, the model sizes of 3D CNNs and ConvLSTM are much larger than those of the general 2D CNNs. When a very deep 3D CNN is used to extract discriminative features from scratch results, the computational cost and memory demand are expensive, which limits the feasibility of real-world applications (\emph{e.g.}, the running time or the hardware requirements). Moreover, ConvLSTM updates the model by backward propagation through time (BPTT), which may result in inefficient training.\footnote{For example, CSRNet~\cite{li2018csrnet} takes $35$ seconds to train one epoch, whereas ConvLSTM~\cite{xiong2017spatiotemporal} spends $530$ seconds training one epoch under the same experimental setting.} Second, the temporal features are encoded at a single temporal scale, while the different regions may require features at different temporal scales. As such, incorporating unnecessary temporal information may deteriorate the performance, which leads to a nonnegligible gap between state-of-the-art image-based methods and existing spatiotemporal methods~(e.g., \cite{zou2019enhanced,xiong2017spatiotemporal,fang2019locality,wu2020triple}). Third, it is time consuming and challenging for humans to annotate large-scale crowds, i.e., annotating dots to the centers of heads, especially for videos. As such, the annotations in videos may not be precise. To deal with imprecise dot annotation, traditional methods obtain a ground truth density map by convolving the annotation dot map with a Gaussian kernel with fixed width and minimize the pixel-wise regression loss between the ground truth density map and predicted density map. However, as shown in Fig.~\ref{Intro_illus}(b), even though two Gaussian blobs are close, the pixel-wise loss is still large, which leads to the unstable training since the ground truth (dot annotation) may not be precise. In contrast, as illustrated in Fig.~\ref{Intro_illus}(a), the proposed PRL stabilizes the training process since it does not enforce a strict alignment of the ground truth and the predicted density map. Moreover, the training curve for the pixel-wise regression loss shows a non-negligible oscillation, which results in a slower convergence rate of the model and may be sensitive to the hyperparameters (\emph{e.g.}, learning rate).

To address these issues, in this paper, we propose a novel SpatioTemporal convolutional Dense network (STDNet) that efficiently extracts spatial and temporal information for crowd counting. Specifically, following P3D-ResNet~\cite{qiu2017learning}, we decouple the Conv3D layer into a Conv2D layer on the spatial dimension and a Conv1D layer on the temporal dimension to address the first issue. Moreover, to address the second issue, we extends the dilated dense convolution to the temporal domain by making the 2D Dense Dilated Convolution Block (DDCB)~\cite{dai2019dense} to be the 3D DDCB, which extracts multiscale temporal features. Moreover, since the multiscale features are extracted to different channels, a general channel attention block is also proposed to attend important channels for different regions. To address the third issue, we propose a patch-wise regression loss (PRL) that calculates the difference between the ground truth density map and predicted density map patch-by-patch.\footnote{It is worth noting that the proposed PRL is general and compatible with both image- and video-based methods. We show the improvement of using PRL with other methods in the experiments.} To demonstrate the improvements of the proposed STDNet, we conduct experiments on the three benchmark datasets UCSD~\cite{chan2008privacy}, Mall~\cite{chen2012feature}, and WorldExpo'10~\cite{zhang2016data,zhang2015cross}, which show that STDNet outperforms the state-of-the-art methods by $10.5$\%, $3.3$\%, and $2$\% in terms of the mean absolute error, respectively.

The main contributions of this paper can be summarized as follows:
\begin{itemize}
\item We propose the SpatioTemporal convolutional Dense Network (STDNet), which alleviates the issues encountered in general 3D CNNs by extending the dilated dense convolution to the temporal dimension with channel attention. The number of parameters of STDNet is only slightly greater than that of image-based methods. To the best of our knowledge, this is the first work that exploits dilated 3D convolution and dense networks in the crowd counting task to deal with video-based datasets.
\item To handle the imprecise annotation of crowds, we propose the patch-wise regression loss (PRL) to mitigate the issues caused by the original pixel-wise regression loss, which stabilizes the training and is compatible with both image- and video-based approaches.
\item Experiments demonstrate the effectiveness of each proposed component, which reveals that STDNet outperforms state-of-the-art models in terms of mean absolute error (MAE) and mean square error (MSE) on the three benchmark datasets. Moreover, the results also validate that the proposed PRL improves the commonly-used pixel-wise regression loss.
\end{itemize}


\begin{figure*}[!t]
\centering
\centerline{\includegraphics[width=\linewidth]{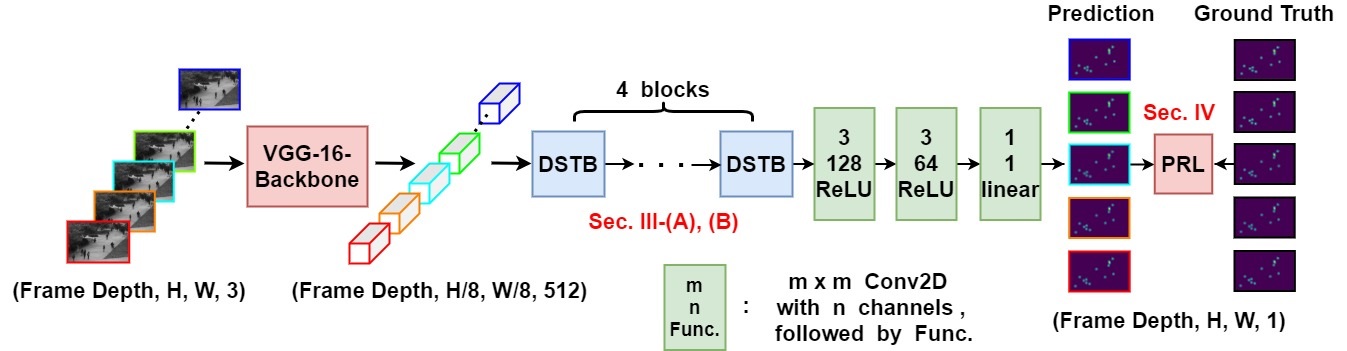}}
\caption{The overall network architecture of our proposed STDNet.}
\label{fig:architecture}
\end{figure*}


\section{Related Works}
\label{Related}
We introduce recent crowd counting works, which can be further categorized into A) detection-based methods, B) regression-based methods and C) density map-based methods. Moreover, we discuss some works that proposed the specialized loss function for the crowd counting task in D) the loss function for crowd counting.

\subsection{Detection-based Methods}
To address the crowd counting task, detection-based methods aim to first detect and locate the head, parts of the body, or the full body for each person by using handcrafted and low-level features~\cite{dollar2011pedestrian,viola2005detecting,brostow2006unsupervised,dalal2005histograms,idrees2013multi,lowe1999object,rabaud2006counting,reg1} and then accumulate the number of corresponding annotations. However, detection is too challenging for extremely crowded scenes due to occlusion and complex patterns, leading to a rapid performance drop. Although there have recently been many advanced deep CNN-based detection methods, e.g., Faster R-CNN~\cite{ren2015faster}, SSD~\cite{liu2016ssd}, YOLO~\cite{redmon2016you}, DecideNet~\cite{liu2018decidenet}, and PSDDN~\cite{liu2019point}, the performance gap between the detection-based methods and the density-map-based methods is still large, especially for densely crowded scenes.

\subsection{Regression-based Methods}
Due to the failure of detection-based methods in extremely congested scenes, there are some studies aiming to directly estimate the total number of people by using machine learning methods, e.g., Gaussian processes~\cite{chan2008privacy}, Bayesian inference~\cite{chan2009bayesian}, and neural networks~\cite{cho1999neural}. However, such methods only consider the relations between the single scalar and the global features of input images while ignoring the informative indication of the local patterns. Moreover, even if the count number is accurate, it provides less information than the density map or detection bounding box since the single scalar cannot infer the location of people.

\subsection{Density-map-based Methods}
The density map, different from the count number, is the label to indicate the existence of humans geometrically, which provides more informative guidance to the training process. Lempisky~\emph{et al.}~\cite{lempitsky2010learning} approximate the counting task to the convex quadratic problem by using the proposed maximum excess over subarray (MESA) loss that optimizes the pixel-level subregion with a high discrepancy to the ground truth. Recently, with the advancement of the deep CNN, a recent line of research has studied the crowd counting problem by CNN-based methods~\cite{cao2018scale,dai2019dense,li2018csrnet,liu2018decidenet,liu2019context,liu2019point,ranjan2018iterative,sam2017switching,sindagi2017generating,tian2019padnet,zhang2016single,jiang2019crowd,shi2019revisiting,yan2019perspective,8767009}. For example, Zhang~\emph{et al.}~\cite{zhang2016single} proposed a multicolumn architecture to extract the multiscale features for solving the variant scale problem. Sam~\emph{et al.}~\cite{sam2017switching} proposed classifying each patch in an image to determine which CNN branch is proper to estimate the number of people. Li~\emph{et al.}~\cite{li2018csrnet} further proposed CSRNet by using the 2D dilated convolutional layer to enlarge the receptive fields without significantly increasing the parameters. Dai~\emph{et al.}~\cite{dai2019dense} proposed the dense dilated convolution block to consider continuously varied scale features. Sindagi~\emph{et al.}~\cite{8767009} proposed two attention modules that enhance low-level features by infusing spatial segmentation information (\textit{e.g.},~\cite{seg1,seg2}) and improved the channel-wise information in the higher level layers. Despite their promising results, most of them put much effort into devising the network with the spatial mechanism without considering the rich information in the temporal domain of video datasets. Few studies~\cite{zou2019enhanced,xiong2017spatiotemporal,fang2019locality,wu2020triple} consider the temporal correlation between consecutive frames. However, the multiscale temporal information is not considered, as well as the imprecise annotations.

\subsection{Loss function for Crowd Counting}
Since the learning target (\emph{i.e.}, the density map) is 2D matrix-form, the intuitive idea is to regress each entry value between the predicted density map and the ground truth, \emph{i.e.}, the pixel-wise regression loss. Although the loss is simple and effective for crowd counting, there are still some studies~\cite{cheng2019learning,liu2018leveraging,ma2019bayesian} that proposed a specialized loss function to avoid the lack of the original loss. For example, Cheng~\emph{et al.}~\cite{cheng2019learning} proposed the maximum excess over pixels (MEP) loss to find the pixel-level subregions that need to be updated to avoid the oversmooth gradients from the pixel-wise regression loss. Liu~\emph{et al.}~\cite{liu2018leveraging} proposed the multitask framework with unlabeled data, using the nature of crowd counting that the number of people in the subimage is definitely less than or equal to that of the corresponding original image, to train the model in a self-supervised manner. Ma~\emph{et al.}~\cite{ma2019bayesian} did not directly regard the density map as the learning target; instead, they used the dot annotations and corresponding Gaussian distribution as the likelihood probability, proposing the Bayesian loss to address the training difficulty from complex and irregular situations (\emph{e.g.}, congestion and occlusion) for crowd counting. Although these works effectively deal with some lack of the original loss, most of them still assume that the given label is accurate. However, it is inconsistent with reality since pixel-level dot annotations are very easily standard-inconsistent and imprecise, and such a problem may lead to inefficient training.


\begin{figure*}[!t]
\centering
\centerline{\includegraphics[width=\linewidth]{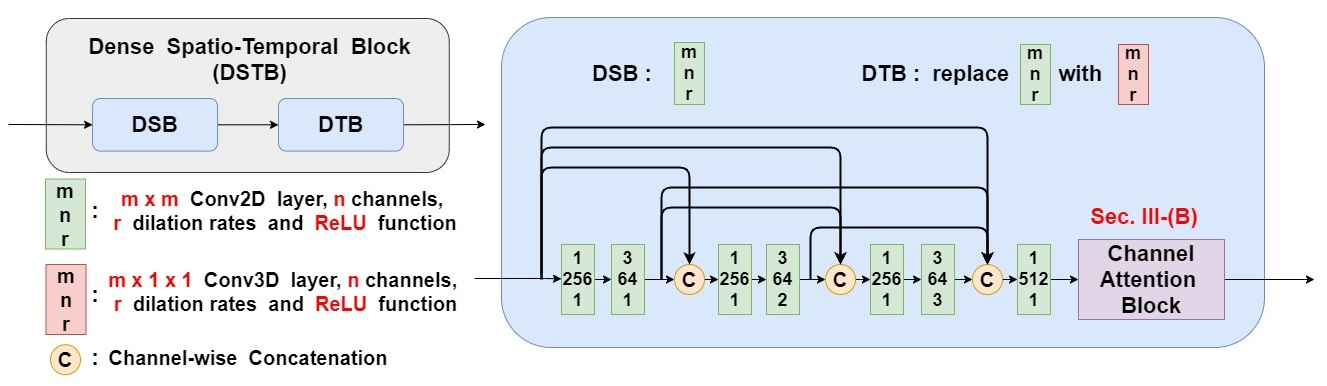}}
\caption{The illustration of the proposed DSTB. It is worth noting that the inner structure of the DTB is similar to DSB, except for the Conv. layer. DSB utilizes Conv2D layer while DTB utilizes the Conv3D layer.}
\label{fig:DSTB}
\end{figure*}


\section{Model architecture}
\label{sec:model architecture}
To capture the multiscale spatiotemporal information for crowd counting without significantly increasing the number of parameters, we propose STDNet, which learns to generate a density map from $T$ consecutive frames. Fig.~\ref{fig:architecture} illustrates the model architecture of STDNet. Specifically, the first ten layers in VGG-16~\cite{simonyan2014very} are used as the backbone to extract features.\footnote{We use the VGG-16 as the backbone for a fair comparison since many baselines use the VGG-16 as the backbone (\emph{e.g.}, CSRNet~\cite{li2018csrnet}, PACNN~\cite{shi2019revisiting}, PaDNet~\cite{tian2019padnet} and CAN~\cite{liu2019context}).} Afterward, we stack four dense spatiotemporal blocks (DSTBs) without any pooling layers in the DSTBs to extract spatiotemporal information while retaining the size of the feature maps. Two Conv2D layers are then stacked after the DSTBs and followed by a$1 \times 1$ Conv2D layer with a $1$ channel to generate the density map. Since there are three pooling layers in the backbone, we simply use bilinear interpolation with a scale of $8$ to upsample the output of the model to match the size of the ground truth. Finally, the predicted density map and the ground truth are fed into the patch-wise regression loss with a gradient-descent-based optimizer to train the parameters. In the following, we first present the proposed DSTB and channel attention block in Sec.~\ref{sec:DSTB} and Sec.~\ref{sec:CAB}, respectively. 
Afterward, the definitions of the ground truth of the density map, geometry-adaptive kernels, and patch-wise regression loss (PRL) are detailed in Sec.~\ref{sec:patch-wise loss}.

\subsection{Dense Spatiotemporal Block}
\label{sec:DSTB}

\subsubsection{Decomposition of 3D CNNs}
To jointly capture the spatial and temporal information without directly using 3D CNNs, inspired by P3D-ResNet~\cite{qiu2017learning}, we decompose a regular 3D CNN to a 2D CNN on the space dimension and a 1D CNN on the temporal dimension. Specifically, given a 3D convolutional filter with the size of $n_{t} \times n_{s1} \times n_{s2}$, where $n_{s1}$ and $n_{s2}$ denote the dimension of the spatial domain and $n_{t}$ denotes the dimension of the temporal domain, we can decompose the 3D convolutional filter to a 2D convolutional filter with the size of $1 \times n_{s1} \times n_{s2}$, which are shared by the spatial domain at different time slots and a 1D convolutional filter with the size of $n_{t} \times 1 \times 1$. The decomposition reduces the number of parameters and takes the temporal features into account simultaneously. For example, when $n_{t}=n_{s1}=n_{s2}=n$, the number of parameters is reduced from $O(n^3)$ to $O(n^2)$ by using the decomposition method.

\subsubsection{Dense Spatiotemporal Block}
The architecture of the dense spatiotemporal block (DSTB) is shown in Fig.~\ref{fig:DSTB}. The DSTB can be decomposed into the dense spatial block (DSB) and the dense temporal block (DTB). The inner structure of the DTB is similar to DSB, except for the size of the kernel, \emph{i.e.},$3\times 3$ for DSB, and $3\times 1\times 1$ for DTB. The key idea
of the dense spatiotemporal block fully exploits the dilated convolution with a dense reception field to capture the spatiotemporal information. Specifically, dilated convolution is regarded as a general and powerful operation for increasing the reception field without any pooling layer or any downsampling layer. As such, we can increase the reception field while retaining the size of the kernel to prevent the model from increasing the number of parameters. Generally, the output of a 3D dilated convolution with a kernel size of $(2L+1)\times(2M+1)\times(2N+1)$ at position $(s,i,j)$, denoted by $y(s,i,j)$, can be calculated as:

\begin{equation}
\label{3D Dilated Convolution}
\centering
\begin{split}
\sum_{l=-L}^{L}\sum_{m=-M}^{M}\sum_{n=-N}^{N} (
w(l,m,n) \cdot x(s+rl,i+rm,j+rn)),
\end{split}
\end{equation}
where $y(\cdot,\cdot,\cdot)$ is the output of the 3D dilated convolutional layer and $x(\cdot,\cdot,\cdot)$ and $w(\cdot,\cdot,\cdot)$ are the input and the convolutional kernel, respectively. The scalar $r$ represents the dilation rate. If $r$ is set to $1$, then Equation~\eqref{3D Dilated Convolution} can be viewed as a regular 3D convolution layer.
The dilation rate controls the reception field in both the temporal and spatial domains, which is helpful for extracting multiscale features.

Inspired by DSNet~\cite{dai2019dense}, we use the same settings for the number of channels and the dilation rates to address the problem of scale variations. Dense channelwise concatenation is also used in the DSTB to present the multiscale features on the channel dimension. Moreover, the $1\times 1$ and $1\times 1\times 1$ Conv. layers are used in front of each dilated Conv. layer to prevent the feature maps from increasing the channel size. It is worth noting that this is the first work combining two mutually beneficial concepts together, i.e., the dense dilated structure provides the representative information on the spatial and temporal domains, while the decomposition trick effectively reduces the number of parameters required to construct a continuous reception field in a dense structure. Moreover, the combination of the decomposition method and the dilated operation can further mitigate the growth of the number of parameters compared with regular 3D convolution.

\begin{figure*}
\centering
\centerline{\includegraphics[width=\linewidth]{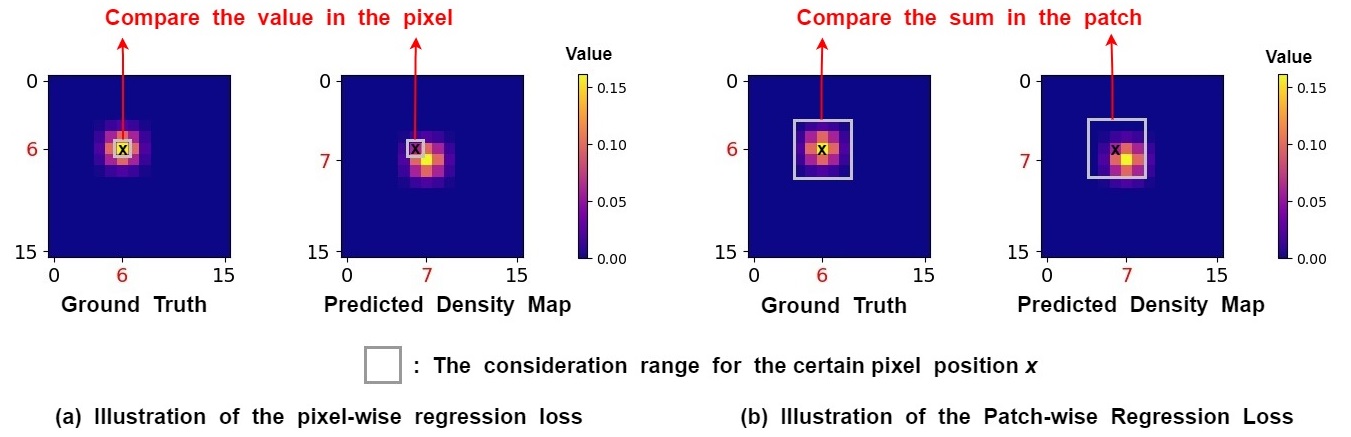}}
\caption{An illustrative example of (a) pixel-wise regression loss and (b) Patch-wise Regression Loss (PRL). The main difference is the considered range for the certain pixel position $\mathbf{x}$: the former is to consider the value only in the pixel for (a), and the latter is to consider the sum of the value in the patch for (b).}
\label{fig:PRL}
\end{figure*}

\subsection{Channel Attention Block}
\label{sec:CAB}
Different from DSNet~\cite{dai2019dense}, we use the general channel attention block inspired by~\cite{zou2019enhanced,hu2018squeeze} to enhance the feature representations for each channel since different regions require features of different scales to predict the correct density map. Therefore, after DSTB extracts the multiscale features by using the different dilation rates and concatenates them on the channel dimension, we further exploit the channel attention block to enhance the extracted features. The general channel attention block can be simply split into the global average pooling layer (GAP) and the fully connected layer. The channel attention blocks in the DSB and the DTB are similar. However, the operation of the GAP layer in the DTB additionally considers the temporal dimension of the feature maps since one of the effects in the DTB is to leverage the temporal information. Specifically, let $C_t$ and $C_s$ denote the number of channels derived from DTB and DSB, respectively. For the input feature maps in the GAP layer, the importance vector $\vec{a}_t = [a_t^{(1)},\cdots,a_t^{(c)},\cdots,a_t^{(C_t)}]$ for the channels in DTB can be calculated by:
\begin{equation}
\label{GAP layer1}
\centering
a_t^{(c)} =\frac{1}{D_{t}\times H_{t}\times W_{t}}\sum_{k=1}^{D_{t}}\sum_{i=1}^{H_{t}}\sum_{j=1}^{W_{t}} x_{t}^{(c)}(k,i,j),
\end{equation}

where $x_{t}^{(c)} \in \mathbb{R}^{D_{t}\times H_{t}\times W_{t}}$ represents the $c$-th channel feature maps derived by DTB. Similarly, the importance vector $\vec{a}_s = [a_s^{(1)},\cdots,a_s^{(c)},\cdots,a_s^{(C_s)}]$ for the channels in DSB is calculated by:

\begin{equation}
\label{GAP layer2}
\centering
a_s^{(c)} =\frac{1}{H_{s}\times W_{s}}\sum_{i=1}^{H_{s}}\sum_{j=1}^{W_{s}} x_{s}^{(c)}(i,j), 
\end{equation}
where $x_{s}^{(c)} \in \mathbb{R}^{H_{s}\times W_{s}}$ represents the feature maps derived by DSB. Following $\vec{a}_t$, we use 2 fully-connected layers to derive the attention vector $\vec{\alpha}_t$, i.e.,  
\begin{equation}
\label{Attention vector in CAB}
\vec{\alpha}_t = Sigmoid\big(W_{2}\times ReLU(W_{1}\times \vec{a}_t)\big),
\end{equation}
where $\vec{\alpha}_t \in \mathbb{R}^{C}$ is the output vector. $W_{1}$ and $W_{2}$ represent the weight matrices of the two fully connected layers. $\vec{\alpha}_s$ can be derived in a similar way. Finally, we use $\vec{\alpha}_t$ and $\vec{\alpha}_s$ to perform channelwise multiplication on the inputs. Since the range of the sigmoid function is defined on $[0,1]$, the channel attention block can adaptively adjust the weights for each channel according to the inputs. Different from DSNet~\cite{dai2019dense}, which leverages dense spatial features, the proposed channel attention block effectively selects the required feature scales for different image regions to address the scale-variant problem. Moreover, the 3D decomposition effectively reduces the required parameters for extracting the spatiotemporal information so that the attention block only has to attend either different temporal scales or different spatial scales.

\section{Patch-wise Regression Loss}
\label{sec:patch-wise loss}
For the objective function, we use the density map as the ground truth and treat crowd counting as a regression problem. The generation of the density map follows previous work~\cite{lempitsky2010learning}. Given the $i$-th ground truth dot map $D_i$, where each dot represents the location of a
person and is marked as 1 on $D_i$, we use a 2D Gaussian kernel to blur $D_i$. The density map of $D_{i}$, denoted as $Y_i^{GT}$, can be defined as:
\begin{equation}
\label{ground truth}
\centering
Y_{i}^{GT}(\mathbf{x}) = \sum_{\{\mathbf{x}'|D(\mathbf{x}')=1\}} \mathcal{N}(\mathbf{x}|\mathbf{x}',\sigma^2I),
\end{equation}
where $\mathbf{x}=(x_1,x_2)$ is the pixel position in the density map, and $\mathcal{N}(\mathbf{x}|\mu, \Sigma)$ is a multivariate Gaussian with mean $\mu$ and covariance $\Sigma$. To address the perspective distortion, MCNN~\cite{zhang2016single} proposed using geometry-adaptive kernels for generating the density map. The definition of geometry-adaptive kernels is similar to Equation~\eqref{ground truth}, except for the standard deviation of the 2D Gaussian kernel, which is formulated as:
\begin{equation}
\label{ground truth with geometry-adaptive kernels}
\centering
Y_{i}^{GT}(\mathbf{x}) = \sum_{\{\mathbf{x}'|D(\mathbf{x}')=1\}} \mathcal{N}(\mathbf{x}|\mathbf{x}',\sigma_{ij}^2I),
\end{equation}
where $\sigma_{ij}$ denotes the standard deviation of the 2D Gaussian kernel for the $j$-th person in $D_{i}$. $\sigma_{ij}$ is defined as $\beta \times \bar{d_{ij}}$, where $\beta$ is a hyperparameter, and $\bar{d_{ij}}$ represents the average distance of $k$ nearest point annotations for the $j$-th person in the $i$-th image.\footnote{For a fair comparison, we set $\beta$ and $k$ as 0.3 and 3, respectively, which follows the setting of MCNN~\cite{zhang2016single}.}

Equipped with the density map, most previous works use the pixel-wise difference between the prediction and the ground truth as the objective loss, which is defined by:
\begin{equation}
\label{pixel-wise l2 loss}
\centering
\mathcal{L}(\mathbf{\theta}) = \frac{1}{2N_b}\sum_{i=1}^{N_b} \left\| f(X_{i};\mathbf{\theta})- Y^{GT}_i\right\|_{2}^{2},
\end{equation}
where $X_i$ represents the $i$-th image, $\theta$ denotes the parameters of the model, $f(\cdot;\theta)$ is the model output, and $N_b$ denotes the total number of images in a batch.\footnote{The factor
$\frac{1}{2}$ is commonly used for gradient-based methods with the $\ell_{2}$-norm.}

However, imprecise (or inconsistent) annotations are very common in densely crowded scenes, especially in videos. Using a pixel-level loss with imprecise annotation may lead to poor performance. For example, if the position of the predicted density peak and that of the ground truth density peak are only one pixel away, which is caused by annotation error, it still induces nonnegligible loss. Therefore, since the semantic patterns for one person are usually filled with several pixels, we propose a patch-wise loss for uncertain matching. Fig.~\ref{fig:PRL}(a) illustrates the original pixel-wise regression loss. Assume that the mean of the Gaussian distribution in the ground truth is $(6,6)$, while the predicted distribution is located at $(7,7)$, to human beings, it is accurate enough to understand the total number of people in the particular region. However, it still induces a considerable error in the pixel-wise loss, which may be harmful to the training process and lead to unstable convergence.

To address the problem, we propose a new loss for crowd counting, i.e., patch-wise regression loss (PRL). The concept of the PRL is illustrated in Fig.~\ref{fig:PRL}(b). Given a pixel position $\mathbf{x}$ in the density map, instead of comparing the value between the prediction and the ground truth, we consider the sum of the density in the patch centered at $\mathbf{x}$. As such, even if the pixel-level error is produced in the training process, it does not lead to a large loss and generates a dominant gradient to update parameters. Moreover, we can use different sizes of patches to consider different levels of precision, e.g., $1\times 1$ (pixel-wise), $3\times 3$, and $5\times 5$. In addition, because the distances to the center pixel $\mathbf{x}$ are different, the importance for each pixel in the patch should be different. Therefore, we use the 2D Gaussian kernel again to imply the weight in the patch. Let $\mathcal{L}_{p}^{(z)}$ and $n_p$ denote the loss of patch size $(2z-1)\times (2z-1)$ and the number of different patch sizes, respectively. The patch-wise regression loss, denoted as $\mathcal{L}_{p}$, is summarized over $n_p$ different patch losses as follows.

\begin{equation}
\label{PRL}
\centering
\mathcal{L}_{p}(\mathbf{\theta}) = \sum_{z=1}^{n_{p}} \lambda_{z}\cdot \mathcal{L}_{p}^{(z)}(\mathbf{\theta}),
\end{equation}
\begin{equation}
\label{PRL-detail}
\centering
\mathcal{L}_{p}^{(z)}(\mathbf{\theta}) = \frac{1}{N_b}\sum_{i=1}^{N_b} \left\|G_{\sigma}^{(z)} \ast f(X_{i};\mathbf{\theta})- G_{\sigma}^{(z)} \ast Y^{GT}_{i}\right\|_{1},
\end{equation}
where $\ast$ denotes the convolution operator, and $G_{\sigma}^{(z)}$ represents the 2D Gaussian kernel with a kernel size of $(2z-1)\times (2z-1)$, e.g., $G_{\sigma}^{(2)}$ is the 2D Gaussian kernel with a kernel size of $3\times 3$. It is worth noting that we use the $\ell_{1}$-norm as our objective loss in Equation~\eqref{PRL-detail} since the $\ell_{2}$-norm loss is sensitive to the outliers from the wrong labeling, which is common to the crowd counting benchmarks.

\section{Experiments}
To evaluate the proposed STDNet, we conduct experiments on three public video-based datasets, including UCSD~\cite{chan2008privacy}, Mall~\cite{chen2012feature}, and WorldExpo'10~\cite{zhang2016data,zhang2015cross} datasets. In the following, we first introduce the evaluation metrics in Sec.~\ref{eval} and present the implementation details in Sec.\ref{imple}. Afterward, the quantitative results on the three datasets with the comparison to the state-of-the-art methods are shown in Sec.~\ref{results}, followed by the qualitative results in Sec.~\ref{sec:qua}. In Sec.~\ref{abla}, we provide the ablation study on UCSD~\cite{chan2008privacy} to demonstrate the improvement of each proposed component. Finally, we discuss the inference speed, compatibility of the proposed PRL, and visualization of channel attention in Sec.~\ref{discuss}.

\subsection{Evaluation Metrics}
\label{eval}
The most common evaluation metrics in crowd counting are the mean absolute error (MAE) and the mean square error (MSE), which are defined as follows.

\begin{equation}
\centering
MAE =\frac{1}{N_t} \sum_{i=1}^{N_t} \left|P_{i} - P_{i}^{GT} \right|,
\end{equation}
\begin{equation}
\centering
MSE = \sqrt{\frac{1}{N_t} \sum_{i=1}^{N_t}  (P_{i} - P_{i}^{GT})^2 },
\end{equation}
where $N_t$ is the total number of images in the testing data, $P_{i}$ and $P_{i}^{GT}$ represent the prediction and the ground truth of the crowd count of the $i$-th image, which are obtained by integrating the predicted density map and the ground truth density map, respectively.

\subsection{Implementation Details}
\label{imple}
Similar to CSRNet~\cite{li2018csrnet}, we use the pretrained weight of the backbone layer (\emph{i.e.}, the first ten layers from VGG-16) to accelerate the training process instead of training from scratch. Adam is utilized as the optimizer~\cite{kingma2014adam} to minimize our proposed $\mathcal{L}_p$, while the learning rate is set to $0.0001$ initially and divided by $2$ per $30$ epochs. Moreover, the ground truth generation settings follow CSRNet~\cite{li2018csrnet}. That is, a fixed kernel with $\sigma = 3$ is set on the UCSD~\cite{chan2008privacy} and the WorldExpo'10~\cite{zhang2016data,zhang2015cross}, while
the deometry-adaptive kernels are used on the Mall~\cite{chen2012feature}. Additionally, we use the horizontal flip to perform the data augmentation. By using cross-validation, the number of patch sizes $n_{p}$ is set to $3$, and the weight $\lambda_{z}$ for each $L^{(z)}(\cdot)$ is set to $1$, $15$ and $3$ for $z=1$, $2$ and $3$.
The standard deviation of the 2D Gaussian kernel defined in Equation~\eqref{PRL-detail} is set to 1. In addition, the frame depths $T$ mentioned in Sec.~\ref{sec:model architecture} on the three datasets (UCSD, Mall, and WorldExpo'10) are set to$10$, $8$, and $5$, respectively, due to the different frame rates of the three datasets. The implementation code will be released on: \url{https://github.com/STDNet/STDNet}.

\setlength{\tabcolsep}{10pt}
\begin{table}[h]
\centering
\caption{Comparisons of different methods on UCSD dataset.}
\label{table:UCSD}
\begin{tabular}{|c|c|c|c|}
\hline
&Method &MAE $\downarrow$ &MSE $\downarrow$\\
\hline
Image-based &RPNet~\cite{yang2020reverse}&1.32&1.23\\
&CSRNet~\cite{li2018csrnet}&1.16&1.47\\
&SPANet+SANet~\cite{cheng2019learning}&1.00&1.28\\
&ADCrowdNet~\cite{liu2019adcrowdnet}&0.98&1.25\\
&PACNN~\cite{shi2019revisiting}&0.89&1.18\\
&PaDNet~\cite{tian2019padnet}&0.85&1.06\\
\hline
Video-based&Bi-ConvLSTM~\cite{xiong2017spatiotemporal}&1.13&1.43\\
&LSTN~\cite{fang2019locality}&1.07&1.39\\
&E3D~\cite{zou2019enhanced}&0.93&1.17\\
\hline
\hline
&STDNet (ours)&\textbf{0.76}&\textbf{1.01}\\
\hline
\end{tabular}
\end{table}

\begin{figure*}[t]
\centering
\centerline{\includegraphics[width=\linewidth]{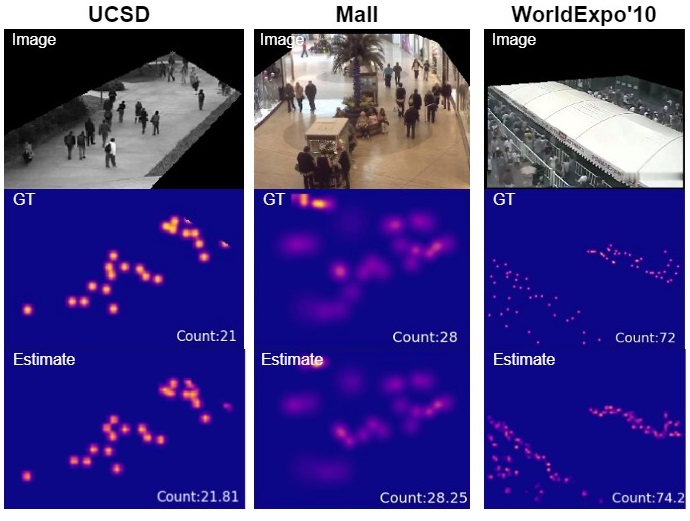}}
\caption{Examples of the predicted results by STDNet on the three datasets, i.e., UCSD, Mall, and WorldExpo'10. The first row shows the input image, the second row represents the ground truth density map, and the last row is the predicted density map.}
\label{fig:Example}
\end{figure*}

\begin{figure*}[h]
\centering
\centerline{\includegraphics[width=\linewidth]{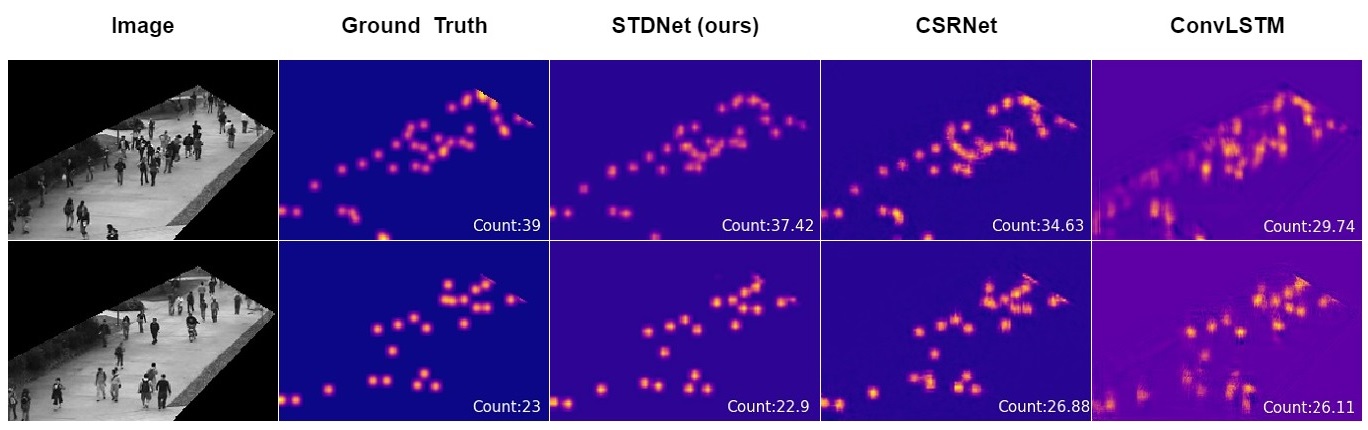}}
\caption{Examples on the UCSD dataset with the comparison to the image- and video-based methods. The first row presents the failure case to all the three models, while the second row presents the success case for the proposed STDNet.}
\label{fig:qualitative results}
\end{figure*}

\subsection{Quantitative Results}
\label{results}
\noindent\textbf{UCSD dataset.} The UCSD dataset~\cite{chan2008privacy} is composed of$2,000$ gray-level frames by a fixed camera in the same scene. According to the settings in CSRNet~\cite{li2018csrnet} and MCNN~\cite{zhang2016single}, the $601$-$1,400$th frames are selected as the training data, and the resting $1,200$ frames are used as the testing set. Table~\ref{table:UCSD} compares the proposed STDNet with the state-of-the-art image- and video-based methods~\cite{cheng2019learning,li2018csrnet,liu2019adcrowdnet,shi2019revisiting,tian2019padnet,xiong2017spatiotemporal,zou2019enhanced,yang2020reverse}. The results show that the proposed STDNet outperforms all of the previous methods in terms of MAE and MSE by at least$10.5$\% and $4.7$\% (compared to PaDNet~\cite{tian2019padnet}), respectively. Additional, the proposed STDNet also outperforms video-based methods, i.e., ConvLSTM~\cite{xiong2017spatiotemporal}, LSTN~\cite{fang2019locality}, and E3D~\cite{zou2019enhanced}, since STDNet effectively extracts the multiscale temporal information by DSTB and further enhances the feature by channel attention.

It is worth noting that the improvement on the UCSD dataset is significant for the following two main reasons. First, the frame rate per second of the UCSD dataset is 10 fps, which is suitable for our work to extract multiscale temporal features since there are high correlations between consecutive frames. For example, if there is a person continuously appearing from the first frame to the last frame, the proposed method can extract the short-term information (\emph{e.g.}, from the 1st frame to the 2nd frame) and the long-term information (\emph{e.g.}, from the 1st frame to the 10th frame). As such, the estimations of both the position and count are improved since the position of the person in each frame should present the monotonic direction, and the count in each frame should remain the same. Second, since the moving speed of every person is usually different, if we can extract multiscale temporal information, all people with different moving speeds can be simultaneously considered to predict better results.

\noindent\textbf{Mall dataset.} The Mall dataset~\cite{chen2012feature} is also a public video-based dataset, which consists of $2,000$ color images in the shopping mall with a resolution of $480\times 640$. For a fair comparison with previous works, the first 800 frames are used as the training set, and the remaining part is used as the testing data.

Table~\ref{table:Mall} compares STDNet with previous works~\cite{chen2012feature,kumagai2017mixture,liu2018decidenet,liu2018crowd,walach2016learning,xiong2017spatiotemporal,zou2019enhanced,zou2020crowd,fang2019locality,wu2020triple}. Similar to the results on the UCSD dataset, STDNet outperforms the image-based methods by at least $3.3$\% in terms of MAE (compared with DecideNet~\cite{liu2018decidenet}). In addition, STDNet also outperforms the four video-based methods~\cite{zou2019enhanced,xiong2017spatiotemporal,fang2019locality,wu2020triple} by at least $10.3$\% in terms of MAE. The results also show that if video-based methods do not leverage the temporal features effectively, the performance is not better than that of the image-based methods.

Note that the frame rate per second of the Mall dataset is 2 fps. Therefore, the improvement benefiting from the temporal information may be smaller. However, we find that the scene in the Mall dataset is more complex than that in the UCSD dataset, \emph{i.e.}, the higher perspective distortion and the occluded situation. Therefore, it is inclined to cause incorrect or imprecise annotations. In this case, the proposed PRL can address this issue since it can match the value between the predicted density map and the ground truth in a patch-by-patch manner, which results in more effective training guidance and outperforms state-of-the-art methods.

\setlength{\tabcolsep}{10pt}
\begin{table}[h]
\centering
\caption{Comparisons of different methods on Mall dataset.}
\label{table:Mall}
\begin{tabular}{|c|c|c|c|}
\hline
&Method &MAE $\downarrow$ &MSE $\downarrow$\\
\hline
Image-based&Chen~\emph{et al.}~\cite{chen2012feature}&3.15&15.7\\
&CNN-Boosting~\cite{walach2016learning}&2.01&-\\
&MoC-CNN~\cite{kumagai2017mixture}&2.75&13.4\\
&HSRNet~\cite{zou2020crowd}&1.80&2.28\\
&DRSAN~\cite{liu2018crowd}&1.72&2.10\\
&DecideNet~\cite{liu2018decidenet}&1.52&1.90\\
\hline
Video-based&Bi-ConvLSTM~\cite{xiong2017spatiotemporal}&2.10&7.6\\
&LSTN~\cite{fang2019locality}&2.00&2.50\\
&TACNN~\cite{wu2020triple}&1.94&2.41\\
&E3D~\cite{zou2019enhanced}&1.64&2.13\\
\hline
\hline
&STDNet (ours)&\textbf{1.47}&\textbf{1.88}\\
\hline
\end{tabular}
\end{table}


\noindent\textbf{WorldExpo'10 dataset.} WorldExpo'10~\cite{zhang2016data,zhang2015cross} contains $3,980$ frames made up of $1,130$ videos in $108$ scenes with a resolution of $576\times 720$. For a fair comparison, $3,380$ frames from $103$ different scenes are used for training, and the remaining $600$ frames from $5$ different scenes are used for testing. Table~\ref{table:worldexpo} compares the proposed STDNet with existing methods~\cite{jiang2019crowd,liu2019context,ranjan2018iterative,shi2019revisiting,xiong2017spatiotemporal,yan2019perspective,zou2019enhanced,yang2020reverse,sam2020locate,miao2020shallow}. The MAE is reported for each scene, and the overall performance is averaged by the results of the different $5$ scenes. From Table.~\ref{table:worldexpo}, the proposed STDNet performs the best in Scene~$1$, Scene~$4$, Scene~$5$ and the average, while ECAN~\cite{liu2019context} and PGCNet~\cite{yan2019perspective} perform the best in Scene~$2$ and Scene~$3$, respectively.


We further investigate Scenes~$2$ and $3$ to find the possible reasons. First, the frame rates per second of the three datasets are different. The frame rate per second of the WorldExpo'10 dataset is $\frac{1}{30}$ only, which is much smaller than that of the UCSD and Mall datasets ($10$ fps and $2$ fps, respectively). Therefore, the frame rate is too low to capture the correlation between the consecutive frames on the WorldExpo'10 dataset, leading to relatively small improvements compared with the state-of-the-art methods. Moreover, we observe that all the video-based approaches~\cite{zou2019enhanced,xiong2017spatiotemporal} perform worse in Scene~$2$ and Scene~$3$ than in others. In fact, Scene~$2$ and Scene~$3$ show more obvious temporal patterns (crowded for watching the performance on the stage and regularly entering the venue in batches, respectively). However, previous frames provide out-of-date information (30 seconds ago) that cannot precisely match the current frame for a better prediction. For example, temporal information shows that people are entering the venue, but a person in the previous frame cannot be found in the current frame after 30 seconds. Since video-based approaches aim to extract the temporal features, the improvement is small if the previous frames provide useless or uncorrelated assistance. It is still worth noting that the average performance of the proposed STDNet is still better than all of the baselines.

 \setlength{\tabcolsep}{5pt}
\begin{table}[h]
\centering
\caption{Comparisons of different methods on WorldExpo'10 dataset.}
\label{table:worldexpo}
\begin{tabular}{|c|c|c|c|c|c|c|}
\hline
Method &S1&S2&S3&S4&S5&Avg.\\
\hline
ic-CNN~\cite{ranjan2018iterative}&17.0&12.3&9.2&8.1&4.7&10.3\\
RPNet~\cite{yang2020reverse}&2.4&10.2&9.7&11.5&3.8&8.2\\
SDANet~\cite{miao2020shallow}&2.0&14.3&12.5&9.5&2.5&8.1\\
PGCNet~\cite{yan2019perspective}&2.5&12.7&\textbf{8.4}&13.7&3.2&8.1\\
TEDnet~\cite{jiang2019crowd}&2.3&10.1&11.3&13.8&2.6&8.0\\
LSC-CNN~\cite{sam2020locate}&2.9&11.3&9.4&12.3&4.3&8.0\\
PACNN~\cite{shi2019revisiting}&2.3&12.5&9.1&11.2&3.8&7.8\\
ECAN~\cite{liu2019context}&2.4&\textbf{9.4}&8.8&11.2&4.0&7.2\\
\hline
Bi-ConvLSTM~\cite{xiong2017spatiotemporal}&6.8&14.5&14.9&13.5&3.1&10.6\\
E3D~\cite{zou2019enhanced}&2.8&12.5&12.9&10.2&3.2&8.32\\
\hline
\hline
STDNet (ours)&\textbf{1.83}&12.78&10.3&\textbf{7.88}&\textbf{2.5}&\textbf{7.05}\\
\hline
\end{tabular}
\end{table}

\begin{figure*}[h]
\centering
\centerline{\includegraphics[width=\linewidth]{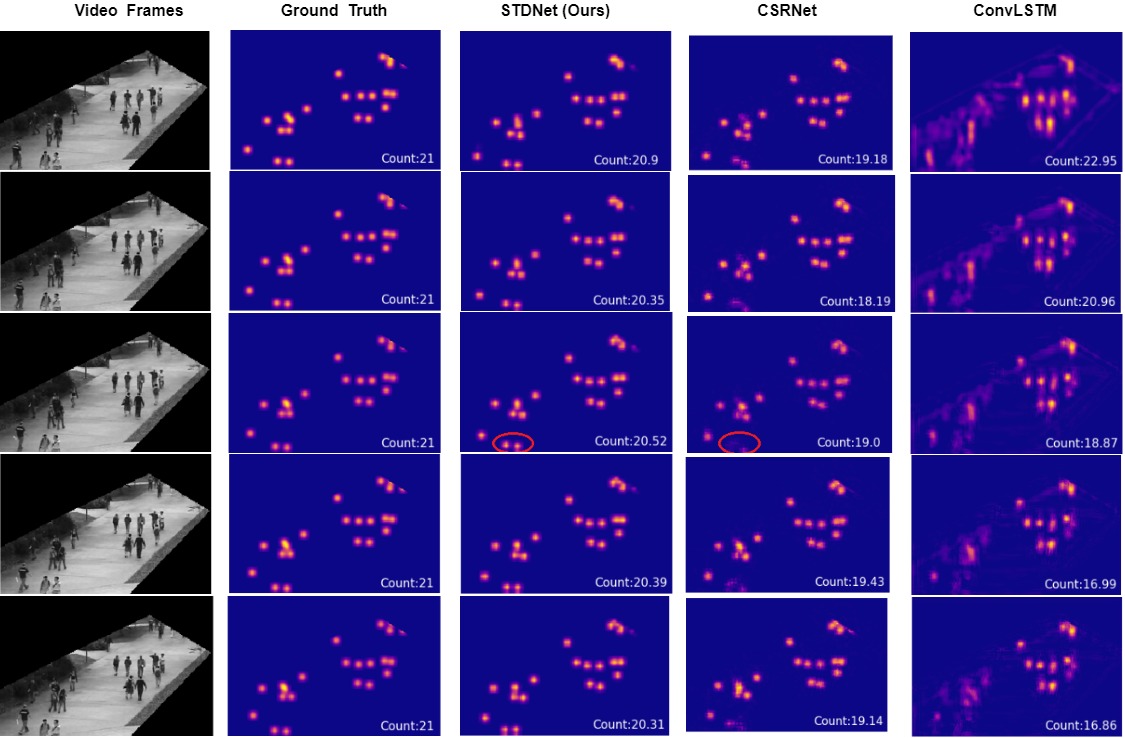}}
\caption{Visual comparison of the image- and video-based methods on consecutive frames of UCSD dataset.}
\label{fig:More Example}
\end{figure*}

\begin{figure*}[h]
\centering
\centerline{\includegraphics[width=\linewidth]{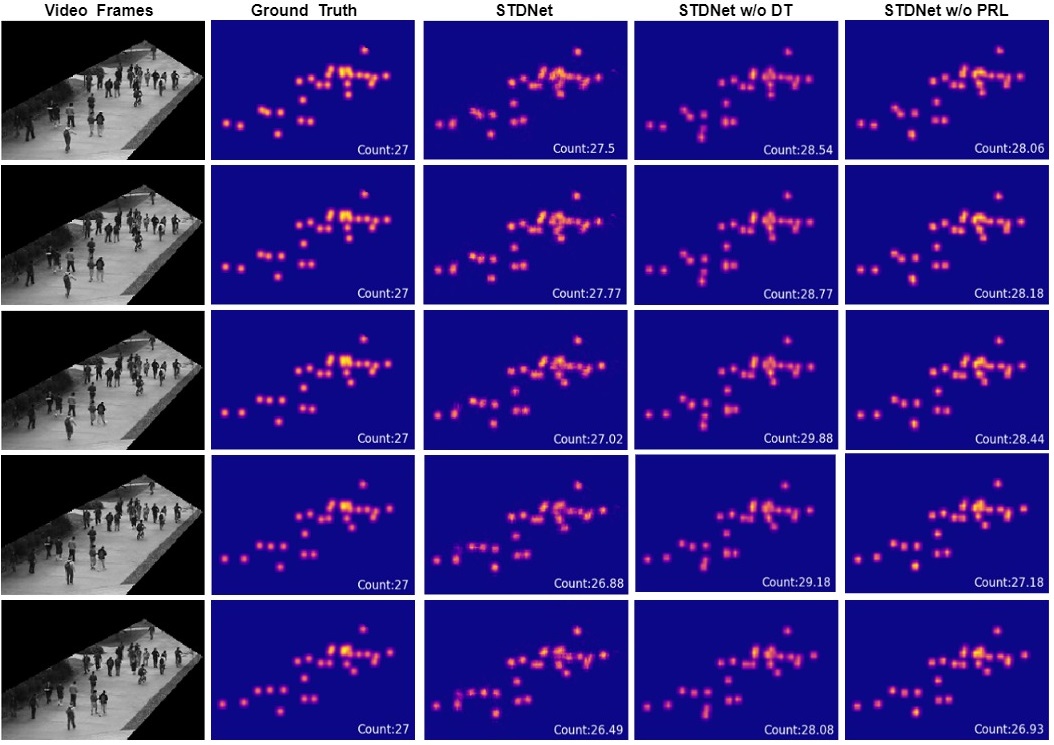}}
\caption{Visual comparison for the ablation studies on consecutive frames of UCSD dataset.}
\label{fig:Abla Example}
\end{figure*}

\begin{figure*}[h]
\centering
\centerline{\includegraphics[width=\linewidth]{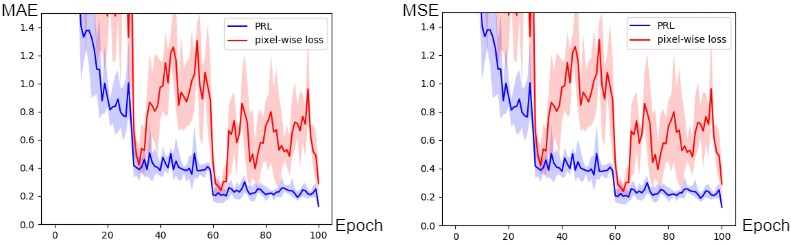}}
\caption{The learning curve of the proposed PRL and the original pixel-wise loss in the training stage in terms of MAE and MSE. The red curve represents the pixel-wise regression loss, while the blue one is the Patch-wise Regression Loss (PRL).}
\label{fig:PRL_learning_curve}
\end{figure*}

\begin{figure*}[h]
\centering
\centerline{\includegraphics[width=\linewidth]{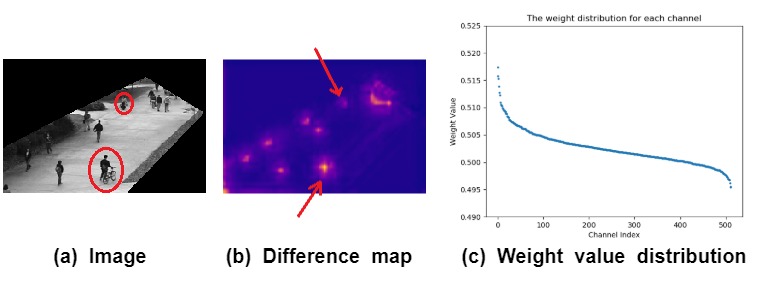}}
\caption{Visualization of the proposed Channel Attention Block in the Dense Spatial Block (Spatial part in the proposed DSTB)}
\label{fig:S_channel_attention}
\end{figure*}

\setlength{\tabcolsep}{6pt}
\begin{table}[t]
\centering
\caption{Leave-one-out ablation analysis with different configurations on the UCSD dataset.}
\label{table:abla}
\begin{tabular}{|c|c|c|c|c|c|c|}
\hline
\multirow{2}*{Method}&\multicolumn{4}{c|}{Configurations}&\multirow{2}*{MAE$\downarrow$}& \multirow{2}*{MSE$\downarrow$}\\
\cline{2-5}
 &CA&DS&DT&PRL&&\\
\hline
\multirow{6}*{Ours}  & & & & &1.25&1.63\\
\cline{2-7}
 & &$\surd$&$\surd$&$\surd$&0.80&1.06\\
\cline{2-7}
 &$\surd$& &$\surd$&$\surd$&0.84&1.07\\
\cline{2-7}
 &$\surd$&$\surd$& &$\surd$&0.83&1.09\\
\cline{2-7}
 &$\surd$&$\surd$&$\surd$& &0.85&1.11\\
\cline{2-7}
 &$\surd$&$\surd$&$\surd$&$\surd$&\textbf{0.76}&\textbf{1.01}\\
\hline
\end{tabular}
\end{table}

In addition, Scene~$4$ is a crowded scene with high perspective distortion, and STDNet still performs the best. It shows that even if the frame rate per second is dramatically distributed on different video-based datasets, our STDNet still achieves state-of-the-art performance due to the proposed DSTB and the channel attention block for extracting and enhancing the different-scale temporal features.

\subsection{Quantitative Results}
\label{sec:qua}

As shown in Fig.~\ref{fig:Example}, the proposed STDNet can not only calculate the crowd count accurately but also precisely estimate the crowd density map. In addition, we also provide some qualitative results compared with the image-based method~\cite{li2018csrnet} and video-based method~\cite{xiong2017spatiotemporal} in Fig.~\ref{fig:qualitative results} (failure case in the first row and success case in the second row). By the success case in Fig.~\ref{fig:qualitative results}, it shows that even if the crowd is sparse with a simple scene, a nonnegligible error still occurs if methods do not carefully extract the information for the prediction. In addition, since our STDNet can adaptively aggregate spatial and temporal features with short-term and long-term information, even in the failure case of Fig.~\ref{fig:qualitative results} with severe occlusion, it still achieves better estimation results compared with other methods.

Fig.~\ref{fig:More Example} visually compares the results of ground truth, STDNet, CSRNet, and ConvLSTM for consecutive frames. The results show that the proposed STDNet better predicts the density maps than CSRNet and ConvLSTM. For example, as shown inside the red circle in the third row and fourth column, CSRNet does not correctly recognize the people, even though it correctly recognize the people in the first frame. This is because CSRNet does not leverage the temporal correlation. Moreover, the results of ConvLSTM are blurry in consecutive frames since ConvLSTM only considers single-scale temporal information while different people may have different walking speeds. In contrast, the proposed STDNet better estimates the density map than the other two methods by extracting the multiscale temporal information and mitigating the inconsistent label issue with PRL.

\subsection{Ablation Study}
\label{abla}


To further investigate the improvement of each proposed component, including the channel attention block (CA), the dense dilated convolution on the spatial (DS) and temporal (DT) dimensions, and the patch-wise regression loss (PRL), we perform a leave-one-out ablation analysis on the UCSD dataset. Table~\ref{table:abla} shows the MAE and MSE with $6$ different configurations. When DS or DT is removed, we use the regular convolutional layer in all the DSBs or DTBs instead of the dilated convolutional layer. Moreover, when PRL is removed, we use the original pixel-wise loss as the objective function.

The results show that MAE and MSE are $1.25$ and $1.63$, respectively, if we remove all the proposed components (the first row of results), while MAE and MSE are improved by 39.2\% and 38.0\%, respectively, if we integrate the four proposed methods into our model (the last row). Moreover, any removal of the proposed module increases the error, which indicates that all four proposed modules are important for video-based crowd counting.

Fig.~\ref{fig:Abla Example} visualizes the qualitative results of the ablation study, i.e., STDNet without the dilated convolution in temporal domain (DT) or without the proposed PRL. The results show that the temporal dilated convolution and the PRL significantly improves the model. This is because the dilated convolution on the temporal domain extracts multiscale temporal features, which improves the model for some complex situations. For example, if an occlusion occur in the current frame, STDNet can estimate the density map more precisely by considering the multiscale features of previous frames. Moreover, the proposed PRL alleviates the issue of imprecise labels to better train the model.


\subsection{Discussion}
\label{discuss}

\subsubsection{Model Size and Processing Speed}
As shown in Table~\ref{table:Model_size_and_training_speed}, we compare STDNet with CSRNet~\cite{li2018csrnet} and ConvLSTM~\cite{xiong2017spatiotemporal} in terms of training time, number of parameters, inference speed, GFLOPS, and MAE on the UCSD dataset. The number of parameters in STDNet is $18.14$ million only, and the training time per epoch is $50$ seconds since we use the decomposition method and the dilated operation. In addition, the inference time of STDNet is $34$ fps. Compared with the image-based CSRNet, our proposed STDNet considers temporal information but only increases the number of parameters by 11.6\%. In contrast, ConvLSTM requires 150.6\% more parameters\footnote{E3D is similar to ConvLSTM but cannot be compared here due to the lack of the design details, \emph{e.g.}, the number of channels in each Conv3D layer.} to deal with temporal information. On the other hand, although the GFLOPS of ConvLSTM is smaller than that of STDNet, the inference speed of ConvLSTM is slower than that of STDNet since the non-linear dependencies between sequence elements of LSTM prevent parallelizing inference over sequence length.

\subsubsection{Generalization and Analysis of PRL}
To demonstrate that the proposed patch-wise regression loss (PRL) is general and compatible with any existing crowd counting methods, including both image- and video-based methods, we compare the results using the original pixel-wise loss and PRL for CSRNet~\cite{li2018csrnet}, CAN~\cite{liu2019context}, and ConvLSTM~\cite{xiong2017spatiotemporal}. As shown in Table~\ref{table:PRL_general}, using the proposed PRL can further improve the results for both image- and video-based methods, including the state-of-the-art ones (\emph{e.g.}, CAN~\cite{liu2019context}). Therefore, the proposed PRL is better than the original pixel-wise loss for crowd counting with imprecise annotations.

\newcommand{\tabincell}[2]{\begin{tabular}{@{}#1@{}}#2\end{tabular}} 
\setlength{\tabcolsep}{2pt}
\begin{table}[t]
\centering
\caption{Comparisons of the training speed, number of parameters, inference time, GFLOPs and MAE on UCSD dataset under the same training settings. Note that the notation ``T'' means the method considers the temporal information.}
\label{table:Model_size_and_training_speed}
\begin{tabular}{|c|c|c|c|c|c|c|}
\hline
Method &T& \tabincell{c}{Training\\time\\(sec./epoch)}& \tabincell{c}{\# para-\\meters}& \tabincell{c}{Inference\\speed\\(fps)}&GFLOPs&MAE\\
\hline
CSRNet~\cite{li2018csrnet}& &35&16.26M&48&2.71&1.16\\
\hline
ConvLSTM~\cite{xiong2017spatiotemporal}&$\surd$&530&40.61M&13&3.27&1.30\\
\hline
STDNet (ours)&$\surd$&50&18.14M&34&4.23&0.76\\
\hline
\end{tabular}
\end{table}

\setlength{\tabcolsep}{6pt}
\begin{table}
\centering
\caption{Comparisons of the proposed Patch-wise Regression Loss (PRL) and the original pixel-wise loss for three methods on UCSD dataset.}
\label{table:PRL_general}
\begin{tabular}{|c|c|c|c|c|c|}
\hline
\multirow{2}*{Method}&\multirow{2}*{ \tabincell{c}{Image- or\\Video-based}}&\multicolumn{2}{c|}{pixel-wise loss}&\multicolumn{2}{c|}{PRL}\\
\cline{3-6}
& &MAE&MSE&MAE&MSE\\
\hline
CSRNet~\cite{li2018csrnet}&Image&1.16&1.47&\textbf{1.06}&\textbf{1.35}\\
\hline
CAN~\cite{liu2019context}&Image&1.08&1.32&\textbf{1.02}&\textbf{1.25}\\
\hline
ConvLSTM~\cite{xiong2017spatiotemporal}&Video&1.30&1.79&\textbf{1.23}&\textbf{1.72}\\
\hline
STDNet (ours)&Video&0.85&1.11&\textbf{0.76}&\textbf{1.01}\\
\hline
\end{tabular}
\end{table}

Moreover, to further analyze the PRL, Fig.~\ref{fig:PRL_learning_curve} shows the training curve in terms of MAE and MSE for the STDNet with PRL and pixel-wise loss. It is worth noting that we repeat the experiments several times to eliminate the randomness, and depict the max, mean and min of the curve. The results show that 1) the proposed PRL (the blue curve) achieves better convergence (solid line) than the original pixel-wise loss (the red curve) for both MAE and MSE, and 2) the oscillation scale (shaded region) of the training curves for the PRL is smaller than that of the pixel-wise loss. This is because it is difficult for the model to precisely estimate the density map in a pixel-level manner and thus induce a large pixel-level difference between the ground truth density map and the predicted density. As such, pixel-wise loss induces a large loss that dominates the gradients and hinders the learning process. In contrast, Fig.~\ref{fig:PRL_learning_curve} shows that our proposed PRL can mitigate the oscillation of the curve and achieve better convergence since we compare the sum of the density value patch-by-patch, avoiding unnecessary guidance from pixel-wise loss; in other words, in the crowd counting task, it is more proper to predict the number of people in a patch-level manner for any method.


\subsubsection{Visualization of Channel Attention Block}
To investigate the effect of channel attention, we visualize the proposed channel attention block. Figs.~\ref{fig:S_channel_attention}(a)-(c) represent the input image, the difference between the input and output of the channel attention block, and the weight value distribution for each channel. It is worth noting that the difference map in Fig.~\ref{fig:S_channel_attention} is derived by 1) adding up the input and output feature maps over the channel dimension and 2) taking the absolute value of the difference between the summation of input feature maps and that of output feature maps. Moreover, in Fig.~\ref{fig:S_channel_attention}(c), we sort the weight values for each channel in descending order for better visualization. In Fig.~\ref{fig:S_channel_attention}(a), we can observe that the scales of human patterns are different (\emph{e.g.}, the two red circles in the image); therefore, by using the proposed channel attention block, it can adaptively highlight the features at different scales for the different regions (\emph{e.g.}, the regions annotated by the two red arrows in Fig.~\ref{fig:S_channel_attention}(b)) to help the model estimate the density map. Fig.~\ref{fig:S_channel_attention}(c) also shows that the channel attention block indeed differentiates the importance of different channels.

\section{Conclusion}
In this paper, we propose a novel spatiotemporal architecture named STDNet for the crowd counting task. STDNet effectively utilizes video data by the proposed spatiotemporal dilated convolution and channel attention block and alleviates the issues generated from the regular 3D CNN or ConvLSTM. Moreover, to handle uncertain or imprecise annotations, PRL is proposed to improve the model. Experimental results on three public datasets show that our model can achieve competitive results compared with other state-of-the-art methods, and the ablation study also demonstrates the effect of each proposed component. In the future, we plan to improve PRL by further exploiting the spatiotemporal information for video-based crowd estimation. Moreover, since the proposed STDNet is compatible with other image-based methods, we will investigate the performance of replacing the spatial convolutional layers with other state-of-the-art image-based methods.

\section{Acknowledgement}
\label{sec:ack} 
This work was supported in part by the Ministry of Science and Technology of Taiwan under Grants MOST-109-2221-E-009-114-MY3, MOST-109-2634-F-009-018, MOST-109-2221-E-001-015, MOST-108-2218-E-002-055, MOST-109-2223-E-009-002-MY3, MOST-109-2218-E-009-025 and MOST-109-2218-E-002-015. We are grateful to the National Center for High-performance Computing for computer time and facilities.


%






\ifCLASSOPTIONcaptionsoff
  \newpage
\fi



%




\bibliographystyle{ieeetr}
\bibliography{reference}

%

\begin{IEEEbiography}[{\includegraphics[width=1in,height=1.25in,clip,keepaspectratio]{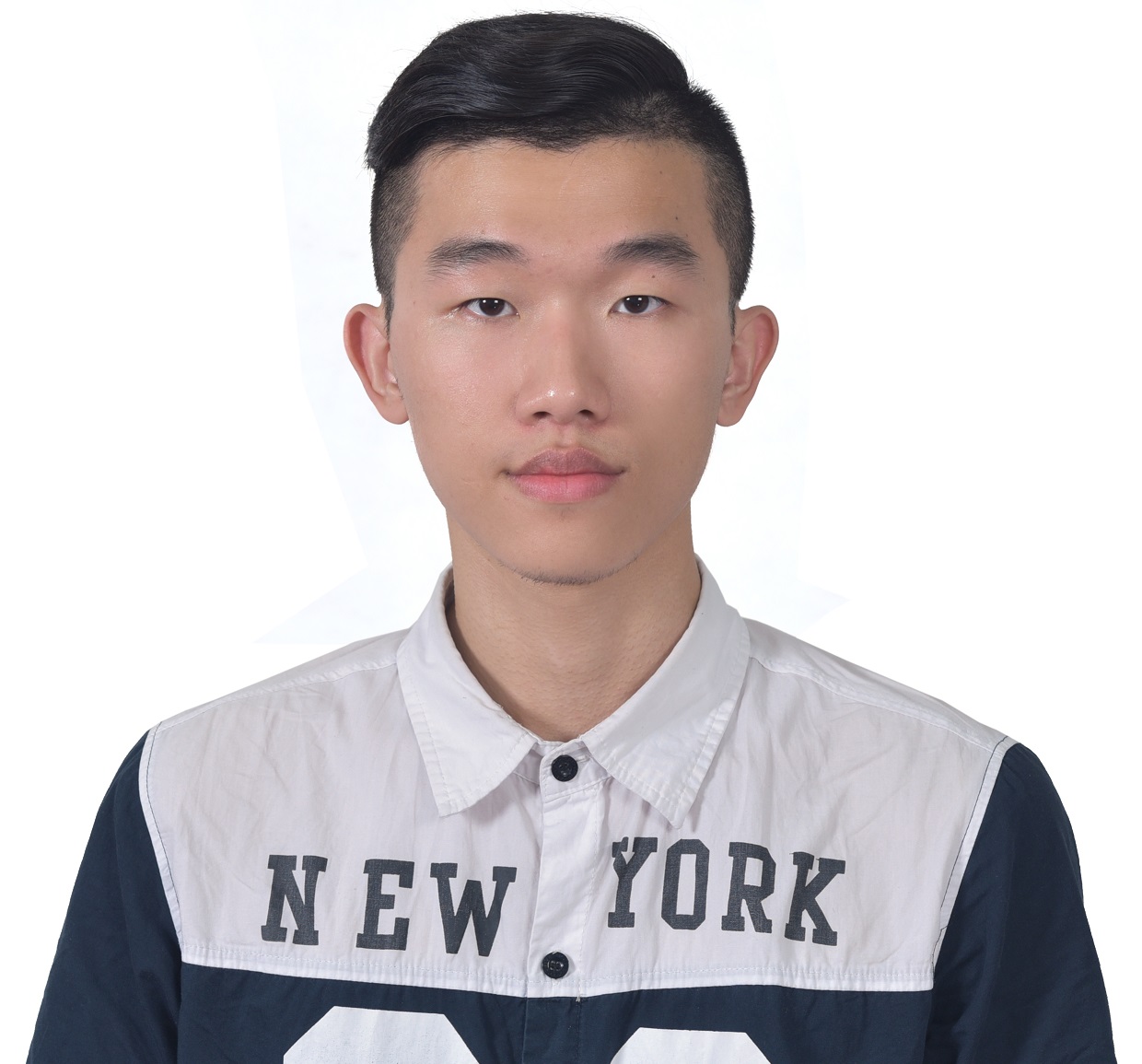}}]{Yu-Jen Ma}
received the B.S. degrees from the Department of Electrical Engineering, National Sun Yat-Sen University (NSYSU), Kaohsiung, Taiwan, R.O.C., in 2019. His research interests include artificial intelligence, deep learning, and computer vision.
\end{IEEEbiography}

\begin{IEEEbiography}[{\includegraphics[width=1in,height=1.25in,clip,keepaspectratio]{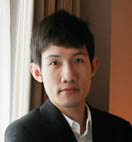}}]{Hong-Han Shuai} received the B.S. degree from the Department of Electrical Engineering, National Taiwan University (NTU), Taipei, Taiwan, R.O.C., in 2007, the M.S. degree in computer science from NTU in 2009, and the Ph.D. degree from Graduate Institute of Communication Engineering, NTU, in 2015. He is now an assistant professor in NCTU. His research interests are in the area of multimedia processing, machine learning, social network analysis, and data mining. His works have appeared in top-tier conferences such as MM, CVPR, AAAI, KDD, WWW, ICDM, CIKM and VLDB, and top-tier journals such as TKDE, TMM and JIOT. Moreover, he has served as the PC member for international conferences including MM, AAAI, IJCAI, WWW, and the invited reviewer for journals including TKDE, TMM, JVCI and JIOT.
\end{IEEEbiography}

\begin{IEEEbiography}[{\includegraphics[width=1in,clip,keepaspectratio]{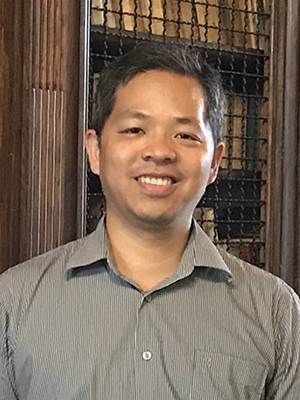}}]{Wen-Huang Cheng} is Professor with the Institute of Electronics, National Chiao Tung University (NCTU), Hsinchu, Taiwan. He is also Jointly Appointed Professor with the Artificial  Intelligence and Data Science Program, National Chung Hsing University (NCHU), Taichung, Taiwan. Before joining NCTU, he led the Multimedia Computing Research Group at the Research Center for Information Technology Innovation (CITI), Academia Sinica, Taipei, Taiwan, from 2010 to 2018. His current research interests include multimedia, artificial intelligence, computer vision, and machine learning. He has actively participated in international events and played important leading roles in prestigious journals and conferences and professional organizations, like Associate Editor for IEEE Transactions on Multimedia, General co-chair for IEEE ICME (2022) and ACM ICMR (2021), Chair-Elect for IEEE MSA technical committee, governing board member for IAPR. He has received numerous research and service awards, including the 2018 MSRA Collaborative Research Award, the 2017 Ta-Yu Wu Memorial Award from Taiwan’s Ministry of Science and Technology (the highest national research honor for young Taiwanese researchers under age 42), the 2017 Significant Research Achievements of Academia Sinica, the Top 10\% Paper Award from the 2015 IEEE MMSP, and the K. T. Li Young Researcher Award from the ACM Taipei/Taiwan Chapter in 2014. He is IET Fellow and ACM Distinguished Member.
\end{IEEEbiography}





\end{document}